\documentclass[a4paper,10pt,twoside]{article}
\usepackage[utf8]{inputenc}

\usepackage{clin}        
\usepackage{harvard}
\usepackage{tikz}
\usepackage{pgfplots}
\usepackage{makecell}
\usepackage{caption}
\usepackage{booktabs}
\usepackage{url}

\pgfplotsset{compat=1.18}

\usepackage[x11names]{xcolor}  
\usetikzlibrary{pgfplots.groupplots}  

\DeclareUnicodeCharacter{0394}{$\Delta$}

\begin{document}

\title{Transcribing Children’s Speech: ASR Performance and Obtaining Reliable Orthographic Transcriptions}

\author{Gus Lathouwers$^*$ \email{guslathouwers@gmail.com}\\
{\normalsize \bf Lingyun Gao}$^*$ \email{lingyun.gao@ru.nl}\\
{\normalsize \bf Catia Cucchiarini}$^*$ \email{catia.cucchiarini@ru.nl}\\
{\normalsize \bf Helmer Strik}$^*$ \email{helmer.strik@ru.nl}\\
\AND \addr{$^*$Radboud University, the Netherlands}}

\maketitle\thispagestyle{empty} 



\begin{abstract}Automatic speech recognition (ASR) has the potential to substantially reduce manual annotation effort in child speech research by generating automatic transcriptions. However, obtaining reliably high-quality ASR transcriptions for child speech remains challenging in low-resource languages due to limited child-specific pre-trained models and highly diverse noise conditions. 
This study investigates the effectiveness of state-of-the-art ASR models on child speech through two research questions, by evaluating nine ASR models from three model families (Whisper, Parakeet, and Wav2Vec2) on two Dutch child speech datasets, JASMIN and DART.
Research question 1 examines the performance of ASR-models applied to child speech. The fine-tuned Whisper-medium model achieves the best overall performance, with a WER of 5.54\% on JASMIN and 70.37\% on DART, showing that the noisy DART data are clearly more challenging.
Research question 2 examines to what extent it is possible to select a subset for which reliable orthographic transcriptions can be obtained automatically, without the need for manual verification. We use an utterance-level selection method that compares ASR output with the original read prompt to identify correctly pronounced recordings. Using the proposed selection method, 42.0\% [for JASMIN] and 18.1\% [for DART] of the utterances can be automatically identified as correctly pronounced with high confidence, resulting in very low error rates on an utterance level (precisions of 98.3\% and higher) and reducing the need for manual verification.
\end{abstract}

\section{Introduction}

Automatic Speech Recognition (ASR) refers to the automatic conversion of speech waveforms into appropriate strings of words \cite{JurafskyMartin2025}. ASR has different use cases, such as automatically transcribing speech in work settings, the annotation of linguistic corpora, and transcription of archival video footage \cite{Russell2024}. As of today, state-of-the-art models show good performance on benchmark datasets across different languages \cite{Yerramreddy2024,Schubert2024}. However, state-of-the-art models may not match performance of trained annotators, especially in cases with high speaker variability or noisy speech quality \cite{Russell2024}. In the case where speech contains a high number of disfluencies or speaker errors, ASR may also struggle much more to produce accurate transcriptions \cite{Alderete2025}.

One focus of ASR has been child speech, as media tools that incorporate ASR have become an important part of classroom learning environments \cite{Bhardwaj2022}. Reliable child ASR, however, has posed challenges. For instance, many popular ASR models, such as Whisper or Wav2Vec2, have overwhelmingly been trained on adult datasets, and do not generalize easily to child speech settings \cite{Jain2023,Asvin2025}. In addition, child-speech recognition in schools is made more difficult due to variable child speaker characteristics \cite{Yadav2021} and difficult recording conditions, such as high levels of background classroom noise \cite{Dutta2022}. ASR-systems have also been found to struggle with verbatim transcriptions, such as those including non-lexical tokens like 'uhm' and 'ehm' \cite{Russell2024}, which can be important to include in child speech transcription for language fluency estimation. To this end, methods like model finetuning on children datasets \cite{Jain2023,Asvin2025,Lileikyte2025}, data-augmentation approaches \cite{Lileikyte2025}, and various linguistic processing techniques \cite{Kathania2021} have been found to increase accuracy. 

For the Dutch language, the use of ASR in child speech may be helpful in reducing human effort in transcribing speech, as in the case of Dutch child reading assessment tools \cite{Groenhof2025,Harmsen2025}. As with child speech ASR for other languages, accurate Dutch child speech ASR faces various challenges like lack of pre-trained models that perform well when applied to Dutch child speech. However, techniques such as finetuning on even low amounts of Dutch child speech, experimental data-augmention, and post-processing normalization or LLM-processing, have all shown to be effective at reducing error rates \cite{Gao2025a,Gao2025b,Shekoufandeh2025,Zhang2024}. Current state-of-the-art model performance depends on the dataset and models used, but may range between 5\% word error rate for low-noise optimal datasets, to 50\% for high-noise datasets \cite{Gao2025a,Zhang2024}. As of yet, there is still a lack of comprehensive evaluation comparing different ASR models under varying noise conditions for Dutch child speech.

Traditionally, metrics such as word error rate are used to benchmark overall ASR model performance. However, such aggregate metrics do not reveal which individual ASR outputs can be considered sufficiently reliable for direct use, particularly in noisy child speech recordings. As a result, a range of approaches have been proposed to assess the reliability of individual ASR predictions at the utterance level, including uncertainty estimators, confidence estimators, and quality estimation methods \cite{Rumberg2023,Li2022,Jalalvand2018}. For child speech, such reliability-oriented approaches are particularly relevant for supporting the automatic transcription of newly collected datasets. While applying ASR models to new data is generally straightforward, determining which of the generated ASR outputs can be accepted with high confidence remains challenging. In this study, a system is formulated that draws on confidence and quality estimation principles. It functions through a selection task that identifies correctly read child speech utterances based on agreement between ASR outputs and the original read prompts, with the aim of reducing the amount of manual verification required to improve the transcriptions.

\subsection{Related Research}

Research on assessing the reliability of ASR output is commonly divided into confidence estimation (CE) and quality estimation (QE) \cite{Negri2014}. Confidence estimation (CE) methods rely on information not contained in the ASR output text itself, such as the original acoustic signal or internal model states, to predict whether a particular ASR output is likely to be correct. In contrast, quality estimation (QE) methods do not require access to the original speech signal or internal model states, and instead assess reliability using only the textual ASR output. Thus, QE systems may analyze textual patterns in an ASR-generated output string, such as repetitions or grammatical inconsistencies, to estimate the likelihood that the transcription is correct or erroneous \cite{Javadi2024}. Conversely, CE approaches require access to the acoustic input or internal model information, such as softmax probabilities, in order to estimate the reliability of an ASR output. 

Confidence estimation (CE) has a long-standing history within ASR, starting with the use of statistical inferential techniques to detect out-of-vocabulary words \cite{Sukkar1996}, and later hybrid ASR systems integrating various sources of information, such as model states and signal features, to determine word-level certainty \cite{Negri2014}. In more recent end-to-end ASR models, token-level softmax probabilities are a straightforward approach for estimating model confidence. However, studies have shown this approach to be unreliable in practice \cite{Kuhn2025,Li2021}. Alternative CE approaches include the use of entropy-loss distributions \cite{Oneata2021}, as well as the integration of auxiliary neural networks trained to estimate the reliability of token-level ASR predictions in end-to-end systems \cite{Li2021}.

Quality estimation (QE) is a comparatively recent direction and has received less attention than confidence estimation (CE) methods. \citeasnoun{Jalalvand2018} proposed a quality estimation approach based on aligning outputs from multiple ASR systems, in which a higher-level decision module selects the most likely correct transcription by comparing features extracted from the different output strings. Subsequently, \citeasnoun{Javadi2024} developed a QE system in which an additional classifier layer is trained on top of a pretrained RoBERTa model. The resulting system, termed NoRefER, predicts transcription error rates using only linguistic features, such as syntax, grammar, and spelling patterns present in the ASR output text. Thus this method bypasses the need for a reference transcription altogether when estimating the WER of an ASR-generated string.

For child speech ASR, relatively little research exists on QE and CE methods. \citeasnoun{Rumberg2023} looked at Connectionist Temporal Classification (CTC) models and found good results in using token-level probabilities to identify errors in Wav2Vec2 output applied to child speech. \citeasnoun{Liu2021} trained a full end-to-end model from scratch on children speech data, and found that attention scores and language model scores, among others, were good predictors for determining reliability of ASR output. Although not strictly a QE method, related work demonstrates that combining outputs from multiple ASR models through LLM prompting can reduce WER, which is conceptually similar to multi-system QE approaches such as \citeasnoun{Jalalvand2018}. In the current study, the task of confidence and quality estimation applied to child speech is approached from a selection-based perspective, focusing on identifying reliable ASR outputs rather than predicting continuous reliability scores. 

\subsection{Current study}

The current research has two aims: 
\begin{enumerate}
    \item This first aim is to establish the performance of state-of-the-art ASR models on Dutch child speech. 
    Because CE, QE, and associated methods of determining reliable speech often depend on the accuracy of the underlying ASR model \cite{Rumberg2023}, we first evaluate baseline performance to determine which models transcribe Dutch child speech most effectively. In total, three ASR-model families, namely Whisper, Wav2Vec2, and Parakeet, will be tested across two datasets. Additionally, because finetuning and prompting have been shown to improve ASR accuracy \cite{Gao2025a,Zhang2024}, we include a prompted Whisper model and fine-tuned variants of all three model families to assess their impact on transcription performance. The first research question is as follows: \\
    \textbf{RQ1: how effective are state-of-the-art ASR-models in transcribing Dutch child speech?}

    \item The second aim of this research is to explore a selection-based approach for automatically identifying reliable ASR outputs. Existing techniques, such as entropy-based loss and token-level softmax scores, have shown varying levels of success applied to speech data, but may not generalize well to the highly variable and noisy speech produced by children \cite{Ravi2025}. For the current research, we use a selection method that marks an ASR output as correct when it exactly matches the corresponding read prompt at the utterance level. This follows the finding that correctly pronounced speech is easier to transcribe than speech that contains speech errors or disfluencies \cite{Alderete2025}. Thus, the inverse should hold true as well, namely that ASR-output that matches a grammatically correct sentence or a correctly spelled word (in this case the original read prompt) should have a higher likelihood of being correct. For our purposes, reliable ASR outputs are defined as outputs selected through utterance-level prompt-matching with low to very low error rate. The research question is: \\
    \textbf{RQ2: To what extent can state-of-the-art ASR-models be used to automatically obtain reliable orthographic transcriptions without manual verification?}
   
\end{enumerate}

\section{Methodology}

In total, three ASR model families (Whisper, Wav2Vec2, Parakeet) were tested across two datasets: JASMIN \cite{cucchiarini2008recording} and the Dutch Automatic Reading Tutor (DART) \cite{Bai2021}. Whereas JASMIN was split, serving both as a training set for fine-tuning and as an evaluation set, DART was used exclusively for evaluation. For an overview of dataset properties, see Table~\ref{tab:dataset_overview}. An overview of all models used is displayed in Table~\ref{tab:asr_models}.

\begin{table}[h!]
\centering
\begin{tabular*}{\textwidth}{@{\extracolsep{\fill}} 
p{0.31\textwidth} 
p{0.23\textwidth} 
p{0.23\textwidth} 
p{0.23\textwidth} 
@{}}
\toprule
\textbf{Property} & \textbf{JASMIN} & \textbf{DART} & \textbf{DART (subset)} \\ 
\midrule
Total speech files (n) 
& 10{,}642 
& 2{,}343 
& 1{,}049 \\

Testing or Evaluation set
& Train / Evaluate 
& Evaluate 
& Evaluate \\

Total Duration
& 9 hours 51 minutes 
& 2 hours 54 minutes
& 0 hours 44 minutes \\

Sentences / Words ratio (\%) 
& 90.3 / 9.7 
& 12.4 / 87.6 
& 8.5 / 91.5 \\

Reading Mistakes Manual (\%)
& 53.5 
& 45.4-53.9 
& 19.8 \\

\bottomrule
\end{tabular*}
\caption{Overview of the datasets used in this study and their key properties. Reading Mistakes Manual (RMM) describes the total amount of utterances in the full datasets in which the children made at least one error pronouncing a prompt.}
\label{tab:dataset_overview}
\end{table}


\subsection{Datasets}
\label{subsec:datasets}

\textbf{JASMIN.} The first dataset used is the  JASMIN-CGN Corpus \cite{cucchiarini2008recording}, an extended version of the Spoken Dutch Corpus \cite{Oostdijk2000}. The full corpus contains speech from different speaker groups, such as the elderly, children, and non native speakers of Dutch. For the current research, only the Dutch speech of native children aged 7-11 was used. The JASMIN corpus contains two types of speech: read speech, in which children read a prompt aloud (typically a short sentence) from a text at their level of reading ability, and dialogue speech, in which children respond spontaneously to machine-generated questions. Each speech utterance in the corpus has a manual orthographic transcription (Manual Orthographic transcription, MO). In addition, for read utterances, the original prompt that the child was instructed to read (PRompt, PR) is also included. The full dataset comprises 9 hours and 51 minutes of speech, of which 7 hours and 50 minutes are read speech and 2 hours and 1 minute are dialogue speech. Among read utterances, 53.5\% contain a pronunciation error, defined as cases where the manual orthographic transcription (MO) differs from the original prompt (PR). 
\noindent
\\
\\
\textbf{DART.} The second dataset used is referenced from the Dutch Automatic Reading Tutor (DART) \cite{Bai2021}. In contrast to the low-noise recordings in JASMIN, the DART corpus contains speech read by children in diverse, noisier environments, with variable microphone quality and background noise. In addition, while JASMIN contains speech of children in different grades, the read speech in DART is from third-graders who are still in the process of learning to read. This implies that they produce broken words, spelled-out words and many other reading errors. The subset of DART used here (n=2343) consists primarily of single words (87.6\%), with the remainder being full sentences, and all utterances were annotated by two independent human annotators. The combined duration of the speech files is 2 hours and 54 minutes. The percentage of utterances containing pronunciation errors ranges from 45.4\% to 53.9\%, depending on the annotation criterion: a maximum of 53.9\% if at least one annotator marked the word as incorrect, and 45.4\% if both annotators did. A separate subset of the DART dataset was also prepared for subsequent analysis (see training and testing procedure below).

\begin{table}[t]
\centering
\begin{tabular*}{\textwidth}{@{\extracolsep{\fill}} 
p{0.24\textwidth}
p{0.46\textwidth}
p{0.08\textwidth}
p{0.10\textwidth}
@{}}
\toprule
\textbf{Model Name} 
& \textbf{Base Model} 
& \textbf{Params (M)} 
& \textbf{Finetuned} \\
\midrule

whisper-large 
& openai/whisper-large-v2  
& 1550 
& - \\

whisper-large-prompted 
& openai/whisper-large-v2 
& 1550 
& - \\

whisper-small-FT 
& openai/whisper-small
& 244 
& v \\

whisper-medium-FT 
& openai/whisper-medium
& 769 
& v \\

whisper-large-FT 
& openai/whisper-large-v2 
& 1550 
& v \\

\midrule

Wav2Vec2-gronlp 
& GroNLP/Wav2Vec2-dutch-large
& 317 
& - \\

Wav2Vec2-FT 
& amsterdamNLP/Wav2Vec2-NL  
& 95 
& v \\

\midrule

parakeet 
& nvidia/parakeet-tdt-0.6b-v3
& 600 
& - \\

parakeet-FT 
& nvidia/parakeet-tdt-0.6b-v3 
& 600 
& v \\

\bottomrule
\end{tabular*}
\caption{Overview of ASR models used in the experiments, including base model, number of parameters, and whether the model was custom finetuned on Dutch child speech in the current study.}
\label{tab:asr_models}
\end{table}

\subsection{Models}

\textbf{Whisper}. Whisper is an encoder–decoder Transformer-based model \cite{Radford2022} that has been extensively used in general speech transcription tasks \cite{Yerramreddy2024} as well as in child speech research \cite{Jain2023,Gao2024}. Applied to Dutch child speech, non-finetuned Whisper WERs have been recorded between 9.4\% and 43\% for JASMIN, with finetuning and custom prompting leading to further reductions in WER \cite{Shekoufandeh2025,Zhang2024,Gao2025b}. Whisper is available in several predefined model sizes that differ in parameter count, including the small (244M parameters), medium (769M parameters), and large (1550M parameters) variants. Though a third version (V3) version of Whisper was released in 2023, we apply Whisper Large-V2 since this was found to suffer less from hallucinations in earlier work \citeasnoun{Gao2025a}. In the current work, the standard, non-prompted, non-fine-tuned Whisper-Large-V2 serves as the baseline model against which other models are evaluated. In addition, we include a prompted Whisper-Large-V2 model derived from a method used by \citeasnoun{Gao2025a}. This approach uses a prompt constructed from a collection of Dutch child speech phrases to promote verbatim transcription, for example, through inclusion of influences and misspelled word transcriptions. See \citeasnoun{Gao2025a} for full details of the prompt and accompanying processing method. In our implementation, we use a lightly modified version of this prompt, with minor token repetitions to further encourage verbatim transcription. We also evaluate three custom fine-tuned Whisper models (small, medium, and large) to assess the effect of model parameter size on performance.
\noindent
\\
\\
\textbf{Wav2Vec2}. Wav2Vec 2.0 is a self-supervised speech representation model that is commonly fine-tuned for ASR using a Connectionist Temporal Classification (CTC) objective, producing token-level predictions aligned to acoustic frames. It has seen widespread use in different ASR applications, such as regular transcription of speech \cite{Yerramreddy2024}, but also as a tool tailored to specific purposes, such as assessing patient intelligibility \cite{Nguyen2024}. Applied to Dutch child speech, WERs of between 13.2 and 24.1\% have been observed, with finetuning of Wav2Vec2 model reducing WER \cite{Fuckner2023,Gao2025b}. For this study, we tested two Wav2Vec2 models with different pretrained bases, one of which was further fine-tuned on Dutch child speech. For the finetuned model, the smaller-parameter (95M) amsterdamNLP/Wav2Vec2-NL was chosen because it was trained exclusively on Dutch speech and is therefore hypothesized to better accommodate the specific speech patterns of Dutch children.
\noindent
\\
\\
\textbf{Parakeet}. NVIDIA Parakeet is a neural transducer–based ASR model using the Recurrent Neural Network Transducer (RNN-T) framework, designed for low-latency and streaming speech recognition. The latest Parakeet model, nvidia-parakeet-tdt-0.6b-v3, was released in August of 2025 and added new support for several languages, including Dutch \cite{Sekoyan2025}. Given their relative novelty, research on their performance on Dutch is non-existent, yet finetuning of the previous Parakeet version model (nvidia-parakeet-tdt-0.6b-v2) in transcribing dysarthric English speech showed good results \cite{Takahashi2025}. For this study, we evaluated the standard, non-finetuned Parakeet v3 model on the two children datasets, alongside a version we fine-tuned on the same Dutch child speech\footnotemark. 

\footnotetext{Nvidia also released a new Canary model at the same time as  Parakeet-tdt-0.6b-v3, namely canary-1b-v2. Here, the Parakeet model was chosen as a target for finetuning over canary-1b-v2 since in preliminary testing the nvidia-parakeet model showed better performance applied to Dutch child speech.}

\subsection{Metrics}

Throughout this paper, terminology related to evaluation of the ASR-output reliability follows \citeasnoun{Molenaar2023}. 
ASR Output (AO) denotes the transcription produced by an ASR model for a given utterance. The read PRompt (PR) is the original text the child was instructed to read in the read condition. 
Manual Orthographic transcription  (MO) refers to the orthographic transcription created by a human annotator. 
Both MO and PR are available for all utterances in the evaluation datasets (JASMIN and DART). For example, JASMIN contains 10,642 speech file utterances, each paired with a corresponding MO and PR. 

For research question 1, ASR Output (AO) is compared to the manual orthographic transcription (MO) to calculate error rates. We then compute Word Error Rate (WER), Character Error Rate (CER), and Utterance Error Rate (UER) for each model. All error rates were calculated through use of the Python package JiWER\footnotemark, which applies Levenshtein distance to align strings. Utterance Error Rate (UER) is the proportion of utterances containing at least one transcription error. For example, UER of 40\% indicates that 60\% of utterances were fully transcribed correctly by a model. For research question 2, UER represents an upper bound on the number of utterances that could potentially be selected as correctly transcribed. 

\footnotetext{Retrieved from \url{https://github.com/jitsi/jiwer}}

For research question 2, the method for identifying correctly transcribed utterances compares ASR Output (AO), the original PRompt (PR), and the Manual Orthographic transcription (MO). For research question 2, all analyses are conducted at the utterance level. For each utterance, we compare the ASR Output (AO) with the original PRompt (PR). This comparison is expressed as a binary value: 1 if the ASR output matches the prompt exactly, and 0 if it does not. We refer to this metric as Reading Mistakes Automatic (RMA). Then, for the same utterance, we compare the PRompt (PR) with the Manual Orthographic transcription (MO). If the child pronounced the prompt correctly, the values match (1); otherwise, they differ (0). This measure is called Reading Mistakes Manual (RMM). Finally, RMA and RMM are compared which yields a confusion matrix that can next be used for classification of true positives (TP), false positives (FP), true negatives (TN), and false negatives (FN). Next, precision (P), and recall (R), and F1 are calculated. 

In the current study, we use AO = PR to identify utterances with a high likelihood of being correct following the observation in other research that speech with few disfluencies and few speech errors is more likely to be transcribed correctly \cite{Alderete2025}. Thus, using AO = PR to select utterances in essence filters out the subset (out of all AO = MO) in which utterances are grammatically sound and spelled correctly. For reference, Table~\ref{tab:transcription_metrics} provides examples of true positive (TP), false positive (FP), true negative (TN), and false negative (FN) cases. 
In addition to precision, recall, and F1, the Matthews Correlation Coefficient (MCC) is calculated for each model, providing a reliable measure even in the presence of class imbalances \cite{Chicco2023}.

\begin{table}[t]
\centering
\begin{tabular*}{\textwidth}{@{\extracolsep{\fill}} 
p{0.21\textwidth}  
p{0.21\textwidth}  
p{0.21\textwidth}  
p{0.07\textwidth}  
p{0.07\textwidth}  
p{0.10\textwidth}  
@{}}
\toprule
\textbf{PR} & \textbf{MO} & \textbf{AO} & \textbf{RMM} & \textbf{RMA} & \textbf{Outcome} \\
\midrule
zij heeft een klacht & zij heeft een klant & we heeft een klans & 0 & 0 & TN \\
duikplank & duikplank & duikplank & 1 & 1 & TP \\
schrapen & schrappen & schrapen & 0 & 1 & FP \\
hoge bomen & hoge bomen & handgebaren & 1 & 0 & FN \\
\bottomrule
\end{tabular*}
\caption{Example comparisons for true positive, true negative, false positive, false negative for research question 2. PR refers to the original prompt the child read, which are included in the JASMIN and DART corpora. MO refers to the manual orthographic transcription, also included in the two corpora used. AO refers to the ASR-prediction generated by a model. RMA (Reading Mistakes Automatic) represents whether AO matches PR in its entirety, if so denoted with a 1. RMM (Reading Mistakes Manual) represents whether MO matches PR in its entirety.}
\label{tab:transcription_metrics}
\end{table}

\subsection{Training and testing procedure}

\textbf{Training/Evaluation.} The JASMIN dataset was used for both model training and evaluation, whereas DART was reserved exclusively for evaluation. For model finetuning, 80\% of the JASMIN dataset (n=8514 of 10642 speech files) was randomly selected for training, with the remaining 20\% (n=2128) used for evaluation. From the n=2128 speech evaluation set, 578 files corresponding to dialogue speech were removed, as these recordings do not contain original read PRompts (PR) required for research question 2. This resulted in a final evaluation set of 1550 utterances. The final set of 1550 utterances was used for both research question 1, which benchmarks ASR model performance, and research question 2, which evaluates methods for selecting reliable ASR output.

For evaluation through DART, two versions of the DART dataset were used: the full dataset (n=2343), containing all original recordings and annotations, and a curated subset (n=1047), in which certain disfluencies and annotation disagreements were removed. Unlike the JASMIN dataset, the full DART dataset (n=2343) included (a) special syntax marking disfluencies (e.g., ‘xxx’ for background noise, ‘*u’ for deviant pronunciations) and (b) a number of utterances for which the two annotators disagreed on whether the original prompt was read correctly. For research question 1, using the full DART set for WER and other error metrics could artificially inflate or deflate model performance. This is because the dataset contains both unannotated background noise and cases where the two annotators disagreed on correctness. Therefore, a curated subset of n=1047 was created by removing utterances containing disfluency markers (e.g., ‘xxx’, ‘*a’) and those with annotator disagreements, providing a cleaner benchmark for model evaluation. For research question 2, both the full DART dataset (n=2343) and the curated subset (n=1047) were used. The curated subset provides a controlled scenario in which background noise and annotation disagreements are removed, while the full dataset represents all original recordings, including noisy and ambiguous cases.

For research question 2, several finetuned models achieved very high precision in selecting correct utterances. This allowed us to explore whether their effects could be combined. In addition to the nine individual models listed in Table~\ref{tab:asr_models}, two model combination strategies were tested. The first strategy, the combined-or-condition (see Table~\ref{tab:classification_results}), counted an utterance as correct (RMA=1) if any of the three finetuned models (whisper-medium-FT, Wav2Vec2-FT, parakeet-FT) produced an ASR Output (AO) that matched the original prompt (PR). This approach effectively leveraged the partial overlap between high-precision models to increase recall while maintaining high overall precision. The second strategy, the combined-and-condition, took a more strict approach to maximize precision. Here, an utterance was classified as correct only if multiple of the three finetuned models produced identical outputs, which then matched the original PR. For the smaller DART dataset, a combined-and-condition strategy in which at least two out of three models agree with each other worked well. For the larger DART dataset (n=2343) and JASMIN, a combined-and-condition in which all three models produced identical output worked best. For the DART subset, the strategy in which two out of three models agreed was deemed more appropriate, as a large number of ambiguous cases had already been filtered out at an earlier stage, reducing the need for stricter agreement (see Section~\ref{subsec:datasets}).

\begin{figure}[!p]
\centering\begin{tikzpicture}
\begin{axis}[
    width=\linewidth,
    height=7cm,
    ylabel={Error Rate (\%)},
    ymin=0, ymax=120,
    xtick={1,2,3,4,5,6,7,8,9},
    xticklabels={
        whisper-large,
        whisper-large-prompted,
        whisper-small-FT,
        whisper-medium-FT,
        whisper-large-FT,
        Wav2Vec2-gronlp,
        Wav2Vec2-FT,
        parakeet,
        parakeet-FT
    },
    x tick label style={rotate=45, anchor=east},
    legend pos=north east,
    title={WER Comparison for JASMIN n=1550} 
]

\addplot+[ color=blue, mark=*, mark options={fill=blue, draw=blue}, solid, nodes near coords, every node near coord/.append style={font=\scriptsize, yshift=2pt}, point meta=explicit ] coordinates {
    (1,16.94) [16.94]
    (2,15.65) [15.65]
    (3,8.94)  [8.94]
    (4,5.54)  [5.54]
    (5,5.46)  [5.46]
};
\addlegendentry{WER}

\addplot+[ color=red, mark=o, mark options={fill=red, draw=red}, solid, nodes near coords, every node near coord/.append style={font=\scriptsize, yshift=2pt}, point meta=explicit ] coordinates {
    (1,58.03) [58.03]
    (2,56.61) [56.61]
    (3,37.72) [37.72]
    (4,27.14) [27.14]
    (5,27.08) [27.08]
};
\addlegendentry{UER}

\addplot+[ color=blue, mark=*, mark options={fill=blue, draw=blue}, solid, nodes near coords, every node near coord/.append style={font=\scriptsize, yshift=2pt}, point meta=explicit ] coordinates {
    (6,23.57) [23.57]
    (7,12.16) [12.16]
};

\addplot+[ color=red, mark=o, mark options={fill=red, draw=red}, solid, nodes near coords, every node near coord/.append style={font=\scriptsize, yshift=2pt}, point meta=explicit ] coordinates {
    (6,75.50) [75.50]
    (7,47.90) [47.90]
};

\addplot+[ color=blue, mark=*, mark options={fill=blue, draw=blue}, solid, nodes near coords, every node near coord/.append style={font=\scriptsize, yshift=2pt}, point meta=explicit ] coordinates {
    (8,17.94) [17.94]
    (9,6.15)  [6.15]
};

\addplot+[ color=red, mark=o, mark options={fill=red, draw=red}, solid, nodes near coords, every node near coord/.append style={font=\scriptsize, yshift=2pt}, point meta=explicit ] coordinates {
    (8,61.64) [61.64]
    (9,29.79) [29.79]
};

\draw[dotted, thick, blue] (axis cs:0,16.94) -- (axis cs:10,16.94);
\node at (axis cs:5.5,16.94) [anchor=south east, font=\small] {Baseline WER};

\end{axis}
\end{tikzpicture}

\centering
\begin{tikzpicture}
\begin{axis}[
    width=\linewidth,
    height=7cm,
    ylabel={Error Rate (\%)},
    ymin=0, ymax=120,
    xtick={1,2,3,4,5,6,7,8,9},
    xticklabels={
        whisper-large,
        whisper-large-prompted,
        whisper-small-FT,
        whisper-medium-FT,
        whisper-large-FT,
        Wav2Vec2-gronlp,
        Wav2Vec2-FT,
        parakeet,
        parakeet-FT
    },
    x tick label style={rotate=45, anchor=east},
    legend pos=north east,
    title={WER Comparison for DART n=1047} 
]

\addplot+[ color=blue, mark=*, mark options={fill=blue, draw=blue}, solid, nodes near coords, every node near coord/.append style={font=\scriptsize, yshift=2pt}, point meta=explicit ] coordinates {
    (1,77.57) [77.57]
    (2,73.37) [73.37]
    (3,75.62) [75.62]
    (4,70.37) [70.37]
    (5,78.40) [78.40]
};
\addlegendentry{WER}

\addplot+[ color=red, mark=o, mark options={fill=red, draw=red}, solid, nodes near coords, every node near coord/.append style={font=\scriptsize, yshift=2pt}, point meta=explicit ] coordinates {
    (1,76.41) [76.41]
    (2,73.45) [73.45]
    (3,70.01) [70.01]
    (4,63.13) [63.13]
    (5,67.72) [67.72]
};
\addlegendentry{UER}

\addplot+[ color=blue, mark=*, mark options={fill=blue, draw=blue}, solid, nodes near coords, every node near coord/.append style={font=\scriptsize, yshift=2pt}, point meta=explicit ] coordinates {
    (6,111.10) [111.10]
    (7,98.68)  [98.68]
};

\addplot+[ color=red, mark=o, mark options={fill=red, draw=red}, solid, nodes near coords, every node near coord/.append style={font=\scriptsize, yshift=2pt}, point meta=explicit ] coordinates {
    (6,94.27) [94.27]
    (7,86.53) [86.53]
};

\addplot+[ color=blue, mark=*, mark options={fill=blue, draw=blue}, solid, nodes near coords, every node near coord/.append style={font=\scriptsize, yshift=2pt}, point meta=explicit ] coordinates {
    (8,99.90) [99.90]
    (9,72.58) [72.58]
};

\addplot+[ color=red, mark=o, mark options={fill=red, draw=red}, solid, nodes near coords, every node near coord/.append style={font=\scriptsize, yshift=2pt}, point meta=explicit ] coordinates {
    (8,93.03) [93.03]
    (9,70.77) [70.77]
};

\draw[dotted, thick, blue] (axis cs:0,77.57) -- (axis cs:10,77.57);
\node at (axis cs:7.5,77.57) [anchor=south east, font=\small] {Baseline WER};

\end{axis}
\end{tikzpicture}
\caption{WER and UER of all models for JASMIN (top graph) and DART (bottom graph). The horizontal dotted line represents the baseline measure (a non-modified version of whisper-large-v2).}
\label{fig:errorrates}
\end{figure}
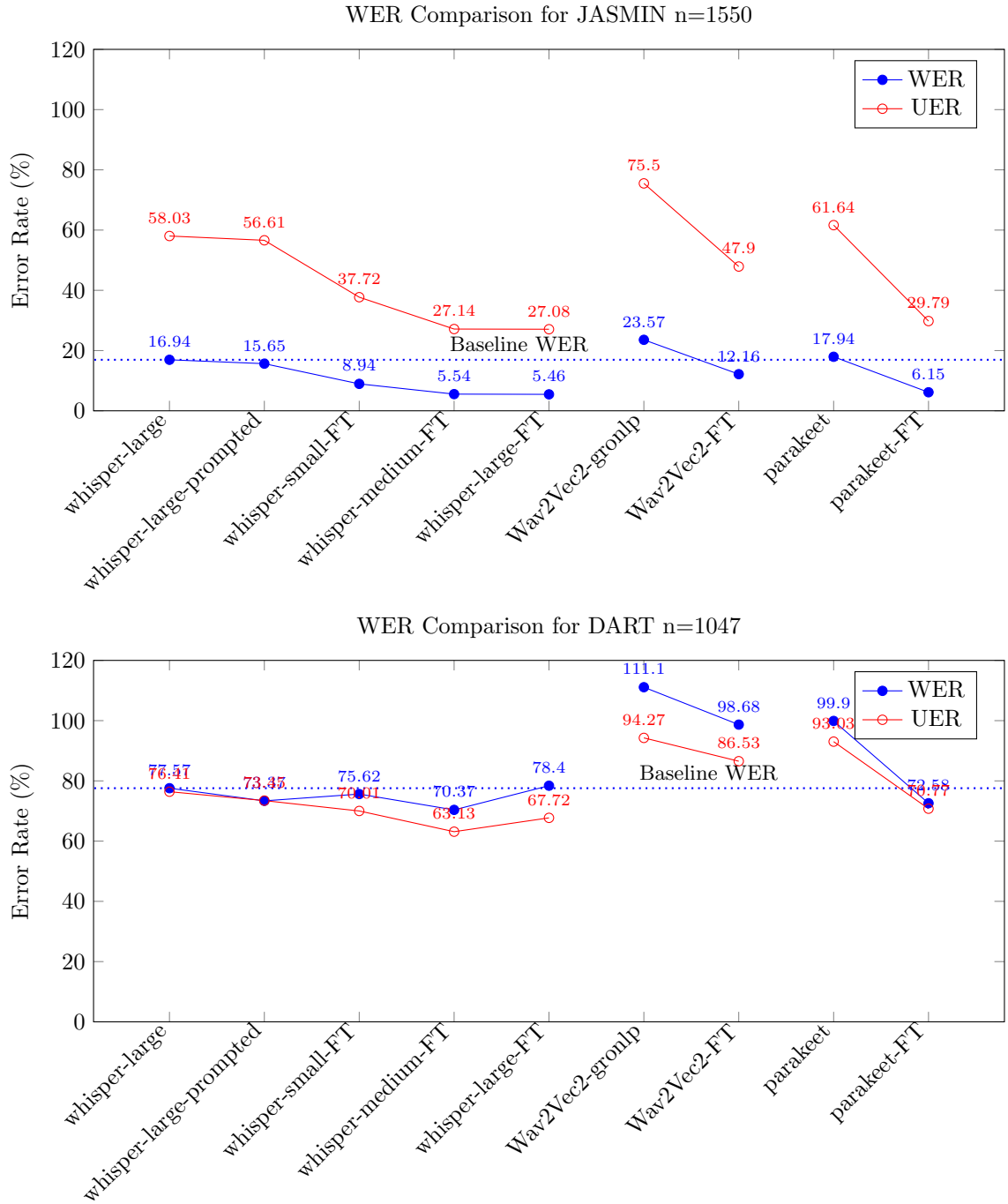

\textbf{Finetuning}. See Table~\ref{tab:asr_models} for an overview of models used and finetuning specifications. All finetuned versions (whisper-small-FT, whisper-medium-FT, whisper-large-FT, Wav2Vec2-FT, parakeet-finetuned-FT) were trained on the same 80\% portion of the JASMIN dataset, containing roughly 8 hours of child speech material\footnotemark. A learning rate of 1e-5 was used for whisper-small-FT, whisper-medium-FT, and whisper-large-FT models. For Wav2Vec2-FT and parakeet-FT, a learning rate of 1e-4 was set. Finetuning for whisper-small-FT model, Wav2Vec2-FT and parakeet-FT was performed on a home GPU (ASUS Dual GeForce RTX 4060 EVO OC Edition 8GB), due to lower amount of total parameters these models have (600 million or fewer). For finetuning whisper-medium-FT and whisper-large-FT, a single A6000 GPU was used. For all models (Whisper, Wav2Vec2, and Parakeet), we performed end-to-end finetuning, updating all parameters without freezing any layers. For the Parakeet model, early testing showed partial layer freezing to adversely affect performance, with full layer training noticeably improving accuracy. All other hyperparameters were kept consistent with those found in the literature \cite{baevski2020,radford2023,Sekoyan2025}

\footnotetext{The train-test split of JASMIN was performed at the utterance level rather than the speaker level, meaning some speakers may have appeared in both splits.}

\section{Results}

Table~\ref{tab:asr_results} and Figure~\ref{fig:errorrates} display results for research question 1. Table~\ref{tab:classification_results} displays results for research question 2.

\begin{table}[t!]
\centering
\begin{tabular*}{\textwidth}{@{\extracolsep{\fill}}
l
ccc
ccc
@{}}
\toprule
& \multicolumn{3}{c}{\textbf{JASMIN (n=1550)}} 
& \multicolumn{3}{c}{\textbf{DART (n=1047)}} \\
\cmidrule(lr){2-4} \cmidrule(lr){5-7}
\textbf{Model} 
& \textbf{WER} & \textbf{CER} & \textbf{UER}
& \textbf{WER} & \textbf{CER} & \textbf{UER} \\
\midrule

whisper-large (baseline) 
& 16.94 & 9.34 & 58.03 
& 77.57 & 45.99 & 76.41 \\

whisper-large-prompted
& 15.65 & 7.11 & 56.61
& 73.37 & 41.29 & 73.45 \\

whisper-small-FT
& 8.94 & 3.33 & 37.72
& 75.62 & 35.44 & 70.01 \\

whisper-medium-FT
& 5.54 & 2.08 & 27.14
& \textbf{70.37} & \textbf{33.27} & \textbf{63.13} \\

whisper-large-FT
& \textbf{5.46} & \textbf{2.03} & \textbf{27.08}
& 78.40 & 38.33 & 67.72 \\

\midrule

Wav2Vec2-gronlp
& 23.57 & 8.35 & 75.50
& 111.10 & 60.75 & 94.27 \\

Wav2Vec2-FT
& 12.16 & 3.74 & 47.90
& 98.68 & 46.00 & 86.53 \\

\midrule

parakeet
& 17.94 & 9.00 & 61.64
& 99.90 & 55.09 & 93.03 \\

parakeet-FT
& 6.15 & 2.37 & 29.79
& 72.58 & 33.81 & 70.77 \\

\bottomrule
\end{tabular*}
\caption{ASR performance on the JASMIN and DART datasets measured in WER, CER, and UER. Bolded values represent the best performance for a metric/dataset combination. The -FT suffix denotes that the model has been finetuned on child audio.}
\label{tab:asr_results}
\end{table}

\subsection{Research question 1: how effective are state-of-the-art ASR-models in transcribing Dutch child speech?}

For the JASMIN dataset (n=1550), ASR models produced WERs ranging from 5.46\% to 23.57\%. The best performing model was whisper-large-FT, with a WER of 5.46\% (CER = 2.03\%, UER = 27.08\%). This is an 11.48\% reduction in WER compared to whisper-large. The worst performance was observed for Wav2Vec2-gronlp, with a WER of 23.57\%. Finetuning reduced WERs across all model types, namely by 11.48\% (whisper-large-FT versus whisper-large), 11.41\% (Wav2Vec2-FT versus Wav2Vec2-gronlp), 11.79\% (parakeet-FT versus parakeet). Prompting also reduced WER by 1.29\% (whisper-large-prompted versus whisper-large).

For the DART dataset, overall WERs were higher, ranging from 70.37\% to 111.10\%. The best result was achieved by whisper-medium-FT, with a WER of 70.37\% (CER = 33.27\%, UER = 63.13\%), a 7.2\% reduction compared to the non-finetuned version of Whisper (whisper-large). As was the case with JASMIN, Wav2Vec2-gronlp was the worst-performing model with a WER of 111.10\%. Finetuning led to WER reductions for all models: 7.2\% (whisper-large-FT versus whisper-large), 12.42\% (Wav2Vec2-FT versus Wav2Vec2-gronlp), and 27.32\% (parakeet-FT versus parakeet). Prompting also improved performance, namely reducing WER by 4.20\%, CER by 4.70\%, and UER by 2.96\%, (whisper-prompted versus whisper-large).

Analysis of the highest-performing model, whisper-medium-FT, shows that the majority of errors were substitutions. See Appendix~\ref{appendix:arq1} for a detailed overview of the most frequent errors.

\subsection{Research question 2: To what extent can state-of-the-art ASR-models be used to automatically obtain reliable orthographic transcriptions?}

\begin{table}[t!]
\centering
\scriptsize
\begin{tabular*}{\textwidth}{@{\extracolsep{\fill}}
l
cccc
cccc
cccc
@{}}
\toprule
& \multicolumn{4}{c}{\textbf{JASMIN (n=1550)}}
& \multicolumn{4}{c}{\textbf{DART (n=1047)}}
& \multicolumn{4}{c}{\textbf{DART (n=2343)}} \\
\cmidrule(lr){2-5}
\cmidrule(lr){6-9}
\cmidrule(lr){10-13}

\textbf{Model}
& \textbf{P} & \textbf{R} & \textbf{F1} & \textbf{MCC}
& \textbf{P} & \textbf{R} & \textbf{F1} & \textbf{MCC}
& \textbf{P} & \textbf{R} & \textbf{F1} & \textbf{MCC} \\
\midrule

whisper-large
& 80.1 & 59.5 & 68.3 & 0.49
& 99.0 & 23.8 & 38.3 & 0.23
& 94.6 & 19.0 & 31.7 & 0.28 \\

whisper-prompted
& 89.2 & 59.7 & 71.7 & 0.59
& 99.6 & 27.3 & 42.9 & 0.25
& 96.2 & 21.6 & 35.2 & 0.31 \\

whisper-small-FT
& 97.9 & 77.8 & 86.7 & 0.79
& 99.6 & 29.6 & 45.6 & 0.27
& 95.4 & 25.6 & 40.3 & 0.32 \\

whisper-medium-FT
& 98.3 & \textbf{88.9} & \textbf{93.4} & \textbf{0.88}
& \textbf{100.0} & \textbf{36.6} & \textbf{53.6} & \textbf{0.32}
& \textbf{99.3} & \textbf{32.9} & \textbf{49.4} & \textbf{0.42} \\

whisper-large-FT
& 96.8 & 89.5 & 93.0 & 0.88
& 99.3 & 31.9 & 48.3 & 0.28
& 97.8 & 27.2 & 42.6 & 0.37 \\

\midrule

Wav2Vec2-gronlp
& 95.7 & 31.3 & 47.2 & 0.42
& \textbf{100.0} & 6.0 & 11.4 & 0.11
& 98.3 & 4.5 & 8.7 & 0.14 \\

Wav2Vec2-FT
& \textbf{98.9} & 64.2 & 77.9 & 0.69
& \textbf{100.0} & 15.5 & 26.8 & 0.18
& 98.1 & 12.0 & 21.4 & 0.23 \\

\midrule

parakeet
& 84.8 & 54.3 & 66.2 & 0.50
& \textbf{100.0} & 7.1 & 13.3 & 0.12
& 90.2 & 5.8 & 10.9 & 0.14 \\

parakeet-FT
& 96.8 & 87.7 & 92.0 & 0.86
& 98.4 & 29.7 & 45.6 & 0.25
& 97.1 & 26.1 & 41.2 & 0.35 \\

\midrule
\midrule

comb-or-condition
& 96.2 & \textbf{94.3} & \textbf{95.2} & \textbf{0.91}
& 99.1 & \textbf{49.2} & \textbf{65.7} & \textbf{0.38}
& 97.9 & \textbf{44.4} & \textbf{61.1} & \textbf{0.50} \\

comb-and-condition$^{\dagger}$
& \textbf{99.3} & 59.4 & 74.4 & 0.66
& \textbf{100.0} & 26.2 & 41.3 & 0.25
& \textbf{100.0} & 5.2 & 9.8 & 0.16 \\

\bottomrule
\end{tabular*}
\caption{Precision (P), Recall (R), F1-score (F1), and Matthews Correlation Coefficient (MCC) for utterance selection across datasets. Best performance for the individual models is displayed in bold (top 9 rows). Conditions in which output of the three finetuned models is combined (whisper-medium-FT, Wav2Vec2-FT, and parkeet-FT) are displayed in the bottom two rows, with the best value for combined strategies also bolded. \\[0.5ex]
$^{\dagger}$For DART n=1047, comb-and-condition agreement is a case in which at least two out of three models of [whisper-medium-FT, Wav2Vec2-FT, parakeet-FT] match the PR. For DART n=2343 and JASMIN n=1550, comb-and-condition are cases in which all three match the PR.}
\label{tab:classification_results}
\end{table}

\textbf{JASMIN}. For the JASMIN dataset, individual ASR-model performance resulted in precision scores ranging from 80.1\% (parakeet) to 98.92\% (Wav2Vec2-FT) and recall scores ranging from 31.3\% (Wav2Vec2-GronLP) to 88.89\% (whisper-medium-FT). 
The best-performing individual model, whisper-medium-FT, achieved P = 98.31\%, R = 88.89\%, F1 = 93.36\% and MCC = 0.88. The P and R numbers are calculated as follows. For 720 (= TP + FN) utterances PR = MO; thus, they are correctly pronounced. 651 (TP + FP) utterances were classified as read correctly by the best model. Of these 651, 640 (TP) were correctly classified, and 11 (FP) were wrongly classified. Then R = 640/720 = 88.89\%, and P = 640/651 = 98.31\%. The total amount of utterances marked as correct, 651 utterances in total, represent 42.0\% of the full dataset (n=1550).

The lowest precision scores were observed for whisper-large (baseline) and parakeet (non-finetuned), with P = 80.11\% and 84.84\%, respectively. Finetuning improved the precision of all three models, namely by 18.2\% for whisper-medium-FT, 3.41\% for Wav2Vec2-FT, and 11.93\% for parakeet-FT. Finetuning also increased recall for all models. Prompting also improved performance, increasing precision by 9.11\% and recall by 0.19\%. For combined model strategies (bottom two rows in Table~\ref{tab:classification_results}), the comb-or-condition resulted in higher recall (R = 94.3\%) but lower precision (P = 96.2\%) than the best individual model. The comb-and-condition achieved higher precision (P = 99.3\%) but lower recall (R = 59.4\%) compared to the best individual model.

\textbf{DART}. For the full DART dataset (n=2343), individual ASR model precisions ranged from 90.2\% to 99.30\%, with recalls between 4.53\% and 32.89\%. As with JASMIN, the best-performing individual ASR model was whisper-medium-FT, achieving P = 99.29\%, R = 32.89\%, F1 = 49.41\%, and MCC = 0.42. Thus, in the full DART dataset, 1280 (= TP + FN) utterances that were correctly pronounced (PR = MO). The whisper-medium-FT model classified 424 (= TP + FP) utterances as correct, of which 421 (TP) were correctly and 3 (FP) were wrongly classified. Following this, R = 421/1280 = 32.89\%, and P = 421/424 = 99.29\%. The total amount of utterances marked as correct, 424 utterances in total, represent 18.1\% of the full dataset (n=2343). Both finetuning and prompting improved P, R, F1, and MCC across all models evaluated.

For the smaller curated DART subset (n = 1047), all models achieved very high precision (98.4\% or higher), while recall scores ranged from 6.0\% (Wav2Vec2-gronlp) to 36.64\% (whisper-medium-FT). For both DART datasets, the comb-or-condition strategy led to increased recall (R = 44.4\% for full DART; R = 49.2\% for curated DART) while maintaining high precision. The comb-and-condition strategy achieved perfect precision (P = 100\%) for both datasets, but with lower recall (R = 5.2\% for full DART; R = 26.2\% for curated DART).

\section{Discussion}

This paper investigated the performance of ASR models on Dutch child speech and a method for selecting correct sentences. The two research questions were: (a) how effective are state-of-the-art ASR-models applied to Dutch child speech, and (b) to what extent can state-of-the-art ASR-models be used to automatically obtain reliable orthographic transcriptions, without manual verification. 

When analyzing the performance of ASR models on Dutch child speech (RQ1), models show different accuracy and error rates applied to the two different Dutch speech datasets. In accordance with earlier research, model error rates applied to a noisy  and more complex dataset (DART) were much higher (WER-range between 70.37\% and 111.10\% in the current study) than when the same ASR models were applied to Dutch child speech recorded in nearly noise-free conditions with fewer reading errors (range between 5.46\% and 23.57\% in the current study) \cite{Bhardwaj2022}. This also mirrors earlier research using JASMIN and DART specifically as evaluation datasets, with DART error rates being notably higher than JASMIN error rates \cite{Gao2025b,Shekoufandeh2025}. Both a custom prompt applied to Whisper, and model finetuning on Dutch child speech, were found to decrease WER for both JASMIN and DART. For JASMIN, the low-noise dataset, the most common errors after finetuning were the result of difficulties in transcribing ambiguous cases, such as when children for example, stumbled in their pronunciations, or transcribing words that have nearly interchangeable solutions (e.g., "er", "d'r") in Dutch. For the noisy dataset with more reading errors, ASR errors stemmed from not correctly distinguishing between vowel patterns and otherwise not registering correctly pronounced words. Additionally, ASR performance on JASMIN in comparison to the DART may be positively biased, as the models were fine-tuned on this dataset and are therefore more likely to adapt to its token distributions and speech patterns.

For identifying correct sentences (RQ2), comparing the ASR output of a model with the original read prompt was shown to be a reliable method. The best-performing individual ASR model achieved a precision of 98.3\% on the dataset it was trained on (JASMIN) and 99.3\% on a noisier dataset (DART). Depending on the dataset, recall ranged from 32.9\% (DART) to 88.9\% (JASMIN), indicating that a substantial proportion of correctly read prompts could be identified using this approach. When applied to full datasets containing both correctly and incorrectly pronounced utterances, 18.1\% of the recordings in DART and 42.0\% in JASMIN could be automatically identified, resulting in very low error rates on utterance level (precisions of 98.3\% and above). Because high-performing ASR models partially differed in which correct utterances they identified, combining model outputs further improved performance by either reducing false positives (comb-and-condition) or increasing the number of utterances selected (comb-or-condition).

To get a better understanding of the methods used and the results obtained, we can look in more detail at the numbers for the best system for the JASMIN data: whisper-medium-FT.
Given that the UER = 27.14\% for this model, then 72.86\% = 1129 out of 1550 utterances are correctly transcribed (AO = MO). UER thus represents an upper bound on the number of utterances that could potentially be selected as correctly transcribed. 
However, if MO is not available, this subset cannot automatically be retrieved.
Therefore, in the method we used to answer RQ2, we proposed to select the utterances for which AO = PR. This subset can automatically be retrieved if MO is not available. As mentioned above, of the 651 (TP + FP) utterances that were selected (classified as read correct) 640 (TP) were correctly classified, and 11 (FP) were wrongly classified. Then R = 640/720 = 98.31\%, and P = 640/651 = 98.31\% for the selected subset.
For the 1550 utterances in total, the proposed method thus automatically obtains a correct orthographic transcription for 640/1550 = 41.29\% of the utterances, and for 11/1550 = 0.71\% an incorrect orthographic transcription was obtained. If one thus wants to obtain orthographic transcriptions for a dataset, this would mean a reduction in human effort of more than 40\% (for this type of data).

The above mechanism through which comparing ASR-output to the intended prompt helps identify low-error utterances may lie in the propensity of ASR-models to be more accurate in transcribing fluent speech instead of speech that contains speech errors \cite{Russell2024}. Thus, a method that identifies utterances with a low number of speech disfluencies or speech errors in some way, essentially marks utterances that have a higher likelihood of being correct. This is further supported by research showing that errors made in speech, such as pronouncing words incorrectly or incompletely, lead to higher WER in ASR-output \cite{Alderete2025}. In the current study, the means for identifying correctly pronounced speech was in matching versus the prompt, but other methods such as for example spelling packages or LLM-processing for detecting errors in syntax or grammar of ASR-output could also potentially be used. Additionally, the current research also aligns with existing research on Dutch child speech, showing that models with WER above 10\% may still be very effective at precise tasks, such as estimating measures of oral fluency \cite{Harmsen2025}. The results from RQ2 here show that models that show a higher error rate (WER of 70.4\%) may still exhibit high precision at selecting reliable ASR-output (P equal to or greater than 99.3\%) on the same dataset.

Comparing the results from RQ1 and RQ2, performance in RQ1 also related to performance in RQ2. For example, the whisper-medium-FT showed a high decrease in WER after finetuning on child speech, which did not translate to an equal proportional drop in WER on the second noisy dataset (70.37\% from 77.57\%, a decrease of 7.2 percentage points), but did cause a rise in precision from 94.6\% to 99.3\%, a 4.7\% increase. Recall was similarly positively affected, also in the smaller curated (DART n=1047) dataset. The same relation also held for other models, where a decrease in WER for JASMIN was positively correlated to increase in precision, Recall, F1, and MCC in RQ2. This is in accordance with existing research showing lower WER may be associated with increases in other downstream task performances, such as assessing intelligibility \cite{Nguyen2024}. Similarly, the method for selecting reliable ASR-output was also found to be reliable for a dataset which contained mostly words (0.12 sentence/word ratio, DART) instead of predominantly sentences (0.90 sentence/word ratio, JASMIN).

Concerning pure individual ASR model performance, the best performing ASR-model in the current study was whisper-medium finetuned on about 8 hours of Dutch child speech data. It slightly outperformed both a finetuned version of a Whisper large model, as well as a finetuned version of Parakeet trained on equal data. It has been observed that the size of training data may require careful matching to model size \cite{Kaplan2020}, and the current result could be ascribed to the 8 hours of data being better fitted to training a medium-size Whisper model (769 million) instead of a large Whisper model (1550 million). Newly tested within the current research is parakeet-tdt-0.6b-v3 model, which has only recently been released \cite{Sekoyan2025}. The v3 of this model added support for the Dutch language and thus allowed finetuning on Dutch child speech. In the current study, the finetuned version of the parakeet-tdt-0.6b-v3 model showed good results, and because of the very high inference speed (up to 10-20 times higher than a regular Whisper model) may be an attractive alternative to Whisper, especially in practical use case applications such as live streaming annotation of child speech \cite{Bhardwaj2022}.

The current study looked at the performance of ASR-models applied to Dutch child speech and a method for automatically selecting correct utterances to reduce the need of human involvement in manually transcribing speech. Limitations of the current paper include the small number of datasets available for testing. Further limitations of the current method used for selecting reliable ASR-output, is that it requires access to the original read prompt, and cannot be used to identify incorrectly pronounced yet correctly transcribed ASR-output. Future research could focus on applying the current method to other languages, or alternatively, include comparisons of ASR model performance applied to adult speech. Given that child read speech is uniquely characterized by speech errors children make because they still are in the process of  learning to read, finding an adult speech dataset that is suitable for comparison is  difficult at the moment. However, new research directions that envisage employing ASR to support low-literate or illiterate adults learning to read could provide relevant and interesting datasets that could be used for this purpose.  Additionally, future research could explore related methods of identifying correct transcriptions that do not make use of an original prompt, especially for speech material types in which the original prompt may not be present. 


\newpage


\bibliographystyle{clin} 
\bibliography{bibliography}


\newpage
\appendix

\section{Research Question 1: Error Analysis}\label{appendix:arq1}

The two tables below display an overview of the most common errors made by the best performing individual model. For the JASMIN, errors were the result of (1) children mispronouncing prompts, (2) children stumbling during pronunciation, (3) very rapid pronunciation of prompts resulting in partial omissions especially in the case of particles, (4) the ASR-model transcribing words that are sometimes interchangeable in written form ("d'r", "er", "haar"), and (5) errors in the manual orthographic transcription done by the original human annotator. The majority of the errors in the JASMIN dataset were substitution errors (71.25\%), with the remainder being deletions (14.52\%) and insertions (14.21\%). For DART, errors were due to (1) model not correctly transcribing vowels (e.g., mixing long and short vowel combinations in dutch such as "o" with "oo"), (2) model dividing a single spoken words into multiple subwords instead of recognizing it as one word (e.g., "de" + "haal" instead of transcribing "verhaal", causing both substitution and insertion errors), (3) low speech quality or background noise creating hallucinations, and (4) ASR-model incorrectly transcribing prompts that were shouted at high volume by children. The majority of the errors in the DART were substitution errors (68.0\%), with the remainder being insertions (30.5\%) and deletions (1.52\%).

\vspace{1em}

\begin{table}[h]
\centering
\begin{tabular*}{\textwidth}{@{\extracolsep{\fill}} 
p{0.08\textwidth} 
p{0.18\textwidth} 
p{0.54\textwidth} 
p{0.16\textwidth} @{}}
\toprule
\textbf{Error} & \textbf{MO $\rightarrow$ AO} & \textbf{Description} & \textbf{Frequency} \\
\midrule

S
& hij $\rightarrow$ haar
& AO transcribes "haar" instead of 'hij' for 3 cases in which child hesitated during pronouncing 'hij'.
& 3 \\

\midrule

S
& een $\rightarrow$ hun
& There are multiple case in which "hun" is transcibed by ASR as "een" and vice versa.
& 3 \\

\midrule

S
& d'r $\rightarrow$ haar
& "d'r" is a short-form common contraction of "haar", in multiple sentences instances of "er", "d'r", and "haar" produce errors between MO and AO.
& 2 \\

\midrule

I
& (none) $\rightarrow$ de
& Insertion of article "de", in two instances the MO forgot to transcribe these.
& 4 \\

\midrule

I
& (none) $\rightarrow$ een
& Insertion of "een", due to hallucination (2 times) or MO error (1 time).
& 3 \\

\midrule

I
& (none) $\rightarrow$ naar
& Insertion of 'naar' in cases in which child spoke very rapidly.
& 2 \\

\midrule

D
& een $\rightarrow$ (none)
& Instances in which a child stumbled during pronouncing words producing several short consonants (e.g., "een la een r man de man rommelt in een la").
& 5 \\

\midrule

D
& het $\rightarrow$ (non)
& Deletion of article "het" due to child pronouncing a previous word unusually, and an instance in which the child only partly pronounced the article "het".
& 2 \\

\bottomrule
\end{tabular*}
\caption{Overview of common errors made for the JASMIN dataset.}
\label{tab:asr_error_overview}
\end{table}

\newpage
\mbox{}

\begin{table}[t!]
\centering
\begin{tabular*}{\textwidth}{@{\extracolsep{\fill}} 
p{0.05\textwidth} 
p{0.21\textwidth} 
p{0.54\textwidth} 
p{0.16\textwidth} @{}}
\toprule
\textbf{Error} & \textbf{MO $\rightarrow$ AO} & \textbf{Description} & \textbf{Frequency} \\
\midrule

S
& pluimpje $\rightarrow$ blijmpje
& Failure to transcribe "plui", instead transcribing "blij" in the word "pluimpje".
& 4 \\
\midrule

S
& besproei $\rightarrow$ bestproe
& Consistently transcribing regularly pronounced 'besproei' as 'bestproe'
& 3 \\
\midrule

S
& fraai $\rightarrow$ vrij
& Three cases, of wich 1 recording has poor speech quality, one fringe case with deviant 'fraai' pronunciation, and one regular case in which ASR-model failed to detect 'raai'.
& 3 \\
\midrule

I
& (none) $\rightarrow$ de
& Many cases in which normally pronounced word was split by ASR into two subwords (e.g., "verhaal" transcribed by ASR-model as "de" + "haa"). 
& 8 \\
\midrule

I
& (none) $\rightarrow$ op 
& Cases in which ASR incorrectly split one spoken word into 2 subwords.
& 6 \\
\midrule

I
& (none) $\rightarrow$ ja 
& A number of cases where caused by the ASR-model transcribing background noise of children yelling, or small partial pronunciations of the word "ja".
& 4 \\
\midrule

D
& een $\rightarrow$ (none)
& Three deletions of "een" in which a child was shouting a prompt repeatedly
& 3 \\

\bottomrule
\end{tabular*}
\caption{Overview of common errors made for the DART dataset.}
\label{tab:asr_error_overview}
\end{table}


\newpage
\clearpage

\section{Research Question 2: Error Analysis}\label{appendix:arq2}

The tables below show all FP errors for the best performing individual model. These resulted from the ASR-model identifying utterances as being correctly pronounced by comparing against PR (full-string ASR-model output matching the original prompt), while these contained errors. For JASMIN, the ASR-model failed to catch small nuances in 11 cases (6 substitution errors, 5 deletion errors), such as missing small consonants or vowel insertions. For DART, out of all utterances the ASR-model identified as being correctly pronounced, three utterances contained an error. All three of these were single words. In one case, the speech was cut-off before being fully pronounced (marked as "*a" by the annotator in "blazen*a"), yet the ASR-model transcribed the full word. In the two other cases, there was unusual pronunciation (drawn out vowel) which the ASR-model did not transcribe.

\vspace{1em}

\begin{table}[h]
\captionsetup{justification=raggedright, singlelinecheck=false}

\centering
\small
\begin{tabular}{p{4.2cm} p{4.2cm} p{0.8cm} p{5cm}}
\toprule
\textbf{Reference (MO)} & \textbf{Hypothesis (AO)} & \textbf{Error} & \textbf{Explanation} \\
\midrule

net als ze zich willen omdraaien blijft de lichtstraal hangen bij een zoort rek &
net als ze zich willen omdraaien blijft de lichtstraal hangen bij een soort rek &
S &
Manual annotator notated “zoort”, ASR-model transcribed “soort”. \\
\midrule

hij w wenkt met zijn arm &
hij * wenkt met zijn arm &
D &
ASR-model deleted extra consonant "w" uttered before "wenkt" in comparison with MO. \\
\midrule

misschien zou ze eigenlijk liever onze hoofden tegen elkaar g aan slaan denkt sarah &
misschien zou ze eigenlijk liever onze hoofden tegen elkaar * aan slaan denkt sarah &
D &
Deleted consonant "g" inserted before "aan slaan". \\
\midrule

robbie schuld zijn hoofd &
robbie schudt zijn hoofd &
S &
“schudt” Instead of "schuld" \\
\midrule

even laten horen ze papa stem &
even later horen ze papas stem &
S &
Missing genitive “’s” for "papa". \\
\midrule

nu zorgt de postbode dat je brieft in de juiste bus terechtkomt &
nu zorgt de postbode dat je brief in de juiste bus terechtkomt &
S&
“brief” Instead of “brieft”. \\
\midrule

maartj denkt na &
maartje denkt na &
S &
Missed the final “e” in “maartje”. \\
\midrule

misschien zijn het wel slangharten &
misschien zijn het wel slangenharten &
S &
Missed syllable “en” within "slangenharten". \\
\midrule

die kan in de in ruimte fel schijnen omdat er geen lucht en wolken zijn &
die kan in de * ruimte fel schijnen omdat er geen lucht en wolken zijn &
D &
Deleted duplicated instance of “in”. \\
\midrule

ze weet niet goed wat ze moet z doen &
ze weet niet goed wat ze moet * doen &
D &
Deletion of letter “z”. \\
\midrule

ik hoop dat je het bij het g rechte eind hebt kerel &
ik hoop dat je het bij het * rechte eind hebt kerel &
D &
Deletion of “g”. \\

\bottomrule
\end{tabular}

\caption{False Positives in the JASMIN dataset (n=1550).}

\medskip
\small

\end{table}

\newpage
\mbox{}

\begin{table}[t!]
\captionsetup{justification=raggedright, singlelinecheck=false}

\centering
\small
\begin{tabular}{p{4.2cm} p{4.2cm} p{0.8cm} p{5cm}}
\toprule
\textbf{Reference (MO)} & \textbf{Hypothesis (AO)} & \textbf{Error} & \textbf{Explanation} \\
\midrule

braden*u &
braden &
S &
Child pronounced the word "braden" incorrectly (marked '*u' by annotator), however ASR-output transcribed "braden". \\\hline

laaaager*u &
lagen &
S &
Drawn out vowel 'a' and 'n' pronounced as 'r' not caught by the ASR-model. \\
\midrule

blazen*a &
blazen &
S &
speech recording was cut-off (marked as '*a' by annotator), ASR-model transcribed full word. \\

\bottomrule
\end{tabular}

\caption{False Positives in the DART dataset (n=2343).}

\medskip
\small

\end{table}

\end{document}